\documentclass[twoside]{article}
\usepackage[accepted]{aistats2018}
\usepackage{amsmath}
\usepackage{algorithm}
\usepackage[noend]{algpseudocode}
\usepackage[psamsfonts]{amssymb}
\usepackage[pdftex]{graphicx}
\usepackage{setspace}
\usepackage{color}
\usepackage{hyperref}

\newcommand{\mbs}[1]{\ensuremath{\boldsymbol{#1}}}
\newcommand{\kv}{\mathbf{k}}
\newcommand{\x}{\mathbf{x}}
\newcommand{\y}{\mathbf{y}}

\newcommand{\secref}[1]{\hyperref[#1]{Section~\ref{#1}}}
\newcommand{\figref}[1]{\hyperref[#1]{Figure~\ref{#1}}}
\newcommand{\algoref}[1]{\hyperref[#1]{Algorithm~\ref{#1}}}

\newcommand{\K}{\mathbf{K}}
\newcommand{\X}{\mathbf{X}}
\newcommand{\N}{\mathbf{N}}

\newcommand{\IG}{\mathcal{IG}}
\newcommand{\cX}{\mathcal{X}}
\newcommand{\RR}{\mathbb{R}}
\newcommand{\me}{\textup{e}}
\newcommand{\Sigmab}{\mbs{\Sigma}}
\newcommand{\ninit}{n_{\text{init}}}

\begin{document}

%

%

\twocolumn[

\aistatstitle{Practical Bayesian optimization in the presence of outliers}

\aistatsauthor{ Ruben Martinez-Cantin \And Kevin Tee \And Michael McCourt }

\aistatsaddress{ SigOpt \\ Centro Universitario De La Defensa \And SigOpt \And SigOpt } ]

\begin{abstract}
  Inference in the presence of outliers is an important field of research as outliers are ubiquitous and may arise across a variety of problems and domains. Bayesian optimization is method that heavily relies on probabilistic inference. This allows outstanding sample efficiency because the probabilistic machinery provides a \emph{memory} of the whole optimization process. However, that virtue becomes a disadvantage when the memory is populated with outliers, inducing bias in the estimation. In this paper, we present an empirical evaluation of Bayesian optimization methods in the presence of outliers. The empirical evidence shows that Bayesian optimization with robust regression often produces suboptimal results. We then propose a new algorithm which combines robust regression (a Gaussian process with Student-$t$ likelihood) with outlier diagnostics to classify data points as outliers or inliers. By using an scheduler for the classification of outliers, our method is more efficient and has better convergence over the standard robust regression. Furthermore, we show that even in controlled situations with no expected outliers, our method is able to produce better results.
\end{abstract}


\section{INTRODUCTION\label{sec:introduction}}
Sample efficient optimization plays an important role in many aspects of science and engineering, where each sample or trial might represent a large cost in time, energy or resources. In recent years, Bayesian optimization has emerged as the \emph{de facto} method for these kind of problems; it provides a black-box solution of the global optimization problem without the need for gradient information \cite{shahriari2016taking}. The underlying mechanism which makes Bayesian optimization methods so powerful is
the use of a probabilistic surrogate model, which incorporates all of the data which has been observed
over the course of the optimization.  This model provides a comprehensive ``memory'' of the
progress of the optimization which empowers Bayesian optimization to often outperform other black-box optimization methods \cite{Moore:1996}.

In the presence of faulty or outlier data, this memory can actually cause problems and slow (or even prevent) convergence because outliers are never forgotten.  Other methods,
such as gradient descent or evolutionary algorithms, have an effectively short memory making them
naturally resilient to outlier data.  For Bayesian optimization methods to manage outliers, steps
must be taken so that they do not hamper the construction of the surrogate model.

Outlier management and detection is an intensive area of research in many disciplines because of the importance of outliers in practice. Outliers are often problem dependent, and are therefore defined differently across different applications. In the context of hyperparameter tuning and design of computer experiments, outliers and gross errors might appear from random bugs, I/O or networking errors, convergence issues for certain sets of parameters, etc. In the case of physical experiments, a user typically has to calibrate the experiment according to the suggested set of parameters and then report the performance, which might result in human mistakes while translating the numbers. Furthermore, in the case of real experiments, external factors might randomly influence and modify certain results. The methods employed to deal with outliers can be classified in two areas: robustness to outliers of inferences and outlier diagnostics \cite{rousseeuw2005robust}.

Robust inference strategies consist of developing models that can incorporate outliers without allowing them to dominate non-outlier data. In the Bayesian framework, we can reduce the influence of the outliers by replacing the \emph{non-robust} population model (e.g., Gaussian) by a longer-tailed distribution, which allows a greater possibility of extreme observations (e.g., Student-$t$) \cite{gelman2014bayesian}. In the Bayesian optimization context, the surrogate model is typically a Gaussian process (GP) \cite{Rasmussen:2006}, which is a regression model from inputs $\x$ to targets $y = f(\x)$. If we consider that the outliers appear in the target variable $y$, we will need to incorporate a long tailed observation model, like a Student-$t$ likelihood. Robust methods are usually more computationally expensive and provide lower accuracy because of the fact that they need to accommodate the long-tailed data. \secref{sec:robust} analyzes in more detail the literature from robust regression, especially as applied to GP regression. Note that errors in the input variables $\x$ are addressed by \emph{sensitivity analysis} \cite{gelman2014bayesian}, which has been already studied in Bayesian optimization \cite{ubonogueira}.

Outlier diagnostics methods generally consist of preprocessing data through statistical analysis to classify outliers and exclude them from a subsequent model built with standard (non-robust) methods. Standard methods such as those found in statistical software are known to be problematic and limited to simpler models \cite{rousseeuw2005robust}. More advanced techniques can be based on statistical learning of outliers.  Supervised learning of outliers requires knowledge of labeled outliers in the same dataset \cite{Hodge2004}. Unsupervised learning requires a good \emph{a priori} model able to represent the data. Details of our diagnostics methodology will be discussed in \secref{sec:diagnostics}.

Methods such as RANSAC, which is popular in computer vision and robotics, combine both strategies: first, a robust model is built accommodating all points, which is then employed for classifying into outliers and inliers. Once the outliers are identified, a non-robust model is used with the inlier data points because it has better statistical properties. Inspired by this methodology, our contribution also includes a two step algorithm that combines robust regression and outlier classification. 



\subsection{Bayesian optimization\label{sec:bayesianoptimization}}
Bayesian optimization methods~\cite{martinez2014bayesopt, shahriari2016taking} try to
minimize, over some domain $\cX\subset\RR^d$, a function $f:\cX\to\RR$ while sampling
$f$ as little as possible (this is what is meant by ``sample efficient'').  The
optimization is generally initialized with $p$ evaluations by sampling with low discrepancy sequences \cite{dick2010digital}
such as latin hypercube sampling.

After the initialization, the sequential component of the optimization begins.
At iteration $t$, all previously observed data $\y=\y_{1:t}$ at points $\X=\X_{1:t}$
is used to construct a probabilistic surrogate model $s_{\y,\X}$.
The next location $\x_{t+1}$ is determined by optimizing a chosen \emph{acquisition function}
which measures the benefit or utility associated with evaluating a proposed $\x\in\cX$.  In this
article, we restrict our focus to only considering expected improvement \cite{Mockus78},
\begin{equation}
	\label{eq:ei}
	EI(\x) = \mathbb{E}_{p(y|s_{\y,\X}(\x))} \left[\max(0,y^* - y)\right],
\end{equation}
where $y^*=\max(y_1,\ldots,y_t)$.

\subsection{Gaussian processes for surrogate modeling\label{sec:gp}}
In \secref{sec:bayesianoptimization} we explained the need for
the probabilistic surrogate model $s_{\y,\X}$ to drive the optimization but not the means
by which it is constructed.
Most frequently, this takes the form of a GP,
although other alternatives
have been presented, such as
random forests~\cite{HutHooLey11-smac},
kernel density estimators~\cite{bergstra2013making} or
Bayesian neural networks~\cite{snoek2015scalable,Springenberg2016}.
For the remainder of the paper we only consider a GP with zero mean and
covariance $k:\cX\times\cX\to\RR$ as the surrogate model.

\paragraph{Observation model} We choose to build our GP models on the belief that data has been observed in the presence of homoscedastic noise $y = f(\x) + \epsilon$, with \emph{Gaussian likelihood} $\epsilon \sim \mathcal{N}(0, \sigma^2_n)$, resulting in a GP posterior model. We can also rewrite the likelihood as $y|f \sim \mathcal{N}(f, \sigma^2_n)$, where $f \equiv f(\x)$. However, as we will see in \secref{sec:robust}, this model is not robust to outliers and we will replace the Gaussian likelihood for a more suitable distribution, that is, the Student-$t$.

In this setting, after $t$ observations (as explained in \secref{sec:bayesianoptimization}), the GP posterior model gives predictions at a query point $\x_q$ which are normally distributed $y_q \sim \mathcal{N}(\mu(\x_q), \sigma^2(\x_q))$, such that
\begin{equation}
  \label{eq:predgp}
  \begin{split}
    \mu(\x_q) &= \kv(\x_q,\X)^T\K^{-1}\y, \\
    \sigma ^2 (\x_q) &= k(\x_q, \x_q) - \kv(\x_q,\X)^T \K^{-1} \kv(\x_q,\X),    
  \end{split}
\end{equation}
where
\begin{equation}
  \label{eq:covmatrix}
  \begin{split}
	\kv(\x_q,\X) &= \begin{pmatrix}k(\x_q,\x_1) &\ldots& k(\x_q,\x_t)\end{pmatrix}^T\!, \\
  \K &= \begin{pmatrix}\kv(\x_1,\X) &\ldots& \kv(\x_t,\X)\end{pmatrix} + \mathbf{I}\sigma^2_n.    
  \end{split}
\end{equation}

The kernel is chosen to be the Mat\'ern kernel with $\nu=5/2$, also called $C^4$ Mat\'ern kernel \cite{fasshauer2015kernel},
\begin{equation}
  \label{eq:kernel}
	k(\x, \x') = \left(1 + r + r^2 / 3\right) \me^{-r},
\end{equation}
where $r=\|\x-\x'\|_{\Lambda}$ for some positive definite matrix $\Lambda$.  The automatic relevance determination kernel which we use here restricts $\Lambda$ to being diagonal. The hyperparameters of $\Lambda$ are estimated by maximum likelihood, although MCMC could be used instead \cite{Snoek2012}.

\paragraph{Contribution}
In this article, we propose a strategy for managing outliers during Bayesian optimization using 
ideas developed in the regression community.
At certain steps during the optimization, we use a GP with the Student-$t$ likelihood to perform an outlier diagnostic.
All previously observed results are then classified as acceptable or outliers and only the acceptable
data is analyzed through the standard Bayesian optimization process.
Our experimental results show that our two-step method for outlier data classification is sufficient for enabling
Bayesian optimization in the presence of outliers. Furthermore, our results also show that this method is preferable to simply using a robust regression model as was previously suggested in \cite{AmarShah2014}, by accommodating the outliers in a robust Bayesian optimization engine. To the authors' knowledge, this is the first work on Bayesian optimization with experimental results addressing the presence of outliers.

\section{ROBUST REGRESSION FOR GAUSSIAN PROCESSES\label{sec:robust}}

\begin{figure*}
	\centering
	\includegraphics[width=0.75\linewidth]{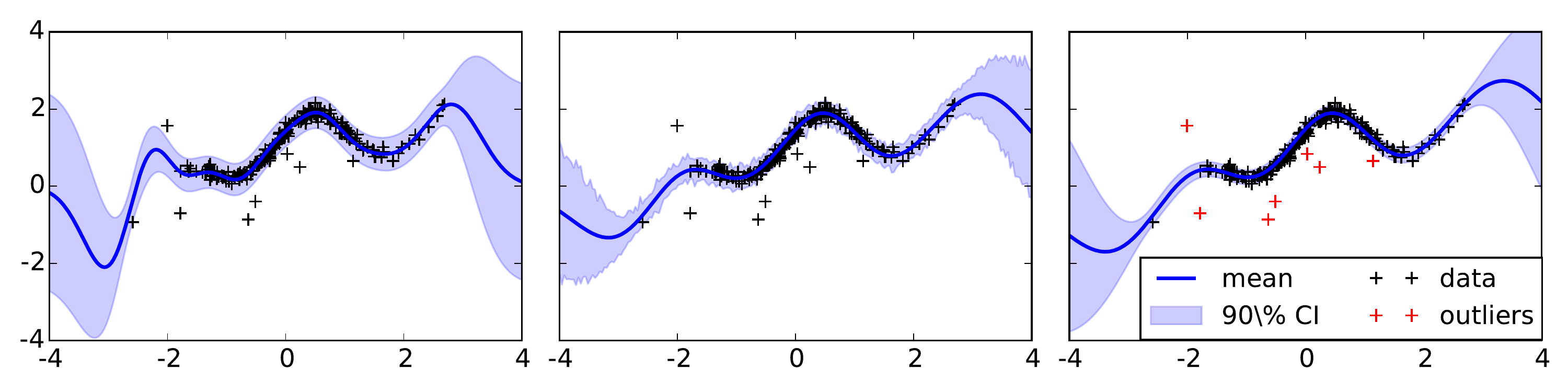}
	\caption{
		\textit{Left}: Regression with outliers using the Gaussian likelihood can yield biased estimates and high variance.
		\textit{Center}: The Student-$t$ likelihood process allows for a better regression, but the estimates with respect to the non-corrupted values is biased and numerically instable.
		\textit{Right}: We use the Student-$t$ likelihood to remove the outliers and use a Gaussian likelihood with the remaining points.  Note how the uncertainty in the left side of the plot is vastly underestimated for the Student-$t$, having the lower bound at $\approx -2$ versus the lower bound of the filtered GP $\approx -4$.
		\label{fig:example}
	}
\end{figure*}

Standard GP-based Bayesian optimization uses an observation model for noisy data with a
Gaussian likelihood, $\epsilon \sim \N(0,\sigma_n^2$) as defined in \eqref{eq:covmatrix}; this allows for closed form inference
but, as shown in \figref{fig:example}, is very sensitive to outliers.
In this section, we review literature on robust regression \cite{rousseeuw2005robust}
to draw inspiration on how to alter the standard GP used
in Bayesian optimization to create a version robust to outliers.

The key change in creating such a regression model is
using a large tail distribution as the observation likelihood in lieu
of the standard Gaussian likelihood;
possible options include the Laplace, the hyperbolic secant, or the Student-$t$ likelihoods.
All those probability distributions are robust to the presence of outliers, with the Student-$t$ likelihood usually providing the best results \cite{jylanki2011robust,lange1989robust}. O'Hagan proved that the Student-$t$ distribution can reject up to $m$ outliers tending to infinity (or negative infinity) provided that there are at least $2m$ observations at all. At the same time, he also showed in \cite{ohagan1979outlier} that the Gaussian distribution is \emph{nonrobust}, meaning that if an outlier is not rejected, the larger the error present in the outlier, the larger the estimate bias will be.

However, the Student-$t$ likelihood, as well as the alternative distributions mentioned, do not allow closed form inference of the posterior.
Therefore, we need to find an approximation that will provide a suitable posterior in the form of a GP or similar.

\paragraph{Related work:} Vanhatalo et al. suggested to use the Laplace approximation to compute the posterior inference of a GP with Student-$t$ likelihood \cite{vanhatalo2009student}. The same authors later compared different strategies: MCMC \cite{neal1997monte}, variational inference, and a modification of the expectation propagation (EP) algorithm with double-loop\cite{jylanki2011robust}. They showed that their modification of the EP is the most robust estimation method, although it has an increased computational cost. It is important to note that the vanilla version of EP does not converge at all for the Student-$t$ likelihood \cite{Rasmussen:2006}. In a different approach, Shah et al. \cite{AmarShah2014} had the surprising result that a Student-$t$ process prior with additive noise in the kernel behaves like a Gaussian or Student-$t$ process posterior with a long-tailed likelihood, similar to the Student-$t$ distribution. The surprise arise for the fact that the Student-$t$ process prior is, by definition, robust to input variables $\x$ but not target variables $y$. The advantage of this method is that it is analytical, removing the extra cost of the iterative approximation. However, the actual statistical properties of the method were unclear. This idea was later proved to have the same marginal likelihood as a Student-$t$ process with dependent Student-$t$ noise, giving a probabilistic interpretation of the results~\cite{Tang2016StudenttPR}. This dependency in the noise might be a strong assumption for certain applications.

Furthermore, the critical parameter controlling robustness of the Student-$t$ distribution it is the degrees of freedom $\nu$, which is recommended to be at least 4 in practice \cite{jylanki2011robust,lange1989robust}. However, this parameter cannot be tuned independently in the dependent case. Furthermore, in the case of noisy data, learning the noise level is harder in the additive noise model due to the entanglement of the variables.
This issue was recently addressed by Tang et al. \cite{ijcai2017-393} by using again the Laplace approximation from Vanhatalo et al. \cite{vanhatalo2009student} to obtain an independent $t$ noise model in a Student-$t$ process.

In the present work, we have decided to compare both approaches: numerical approximation of the Student-$t$ likelihood and the use of a Student-$t$ process with additive noise.

\subsection{Numerical approximation of Student-$t$ likelihood\label{sec:laplace}}
 First, we will use the Student-$t$ likelihood from Vanhatalo et al. \cite{vanhatalo2009student}. The Student-$t$ distribution has the form
\begin{equation}
	\label{eq:likt}
	t(y;f,\sigma_0,\nu) = \frac{\Gamma\left(\nu + \frac{1}{2}\right)}{\Gamma(\frac{\nu}{2}) \sqrt{(\nu\pi)\sigma_0}}\left[1+\frac{(y-f)^2}{\nu\sigma_0^2}\right]^{-\nu - \frac{1}{2}},
\end{equation}
where $f \equiv f(\x)$, $\nu$ is the degrees of freedom and $\sigma_0$ is the scale parameter. In the Bayesian context, the Student-$t$ distribution usually arises from a normal distribution with a conjugate hyperprior on the variance variable, such as the inverse-$\chi^2$, the inverse gamma or even the Jeffreys prior \cite{Santner03}. For example, in this case, the model $y|f \sim t(y;f,\sigma_0,\nu)$  is equivalent to the original Gaussian likelihood with an hyperprior on the noise term $\sigma_n$. That is:
\begin{equation}
  \label{eq:likthierarchical}
  \begin{split}
    y | f, \sigma_n &\sim \mathcal{N}(f, \sigma^2_n)\\
    \sigma_n^2 | \nu, \sigma_0^2 &\sim \chi^{-2}(\nu, \sigma_0^2)
  \end{split}
\end{equation}

Jylanki et at. \cite{jylanki2011robust} present different approximation methods for which we implemented the Laplace method (the simplest and most extended method) \cite{vanhatalo2009student}.
These works were mostly intended for regression applications where large amounts of data are available at once. In contrast, Bayesian optimization seeks to minimize the number of data points, often resulting in less data than most regression applications. Furthermore, observations arrive sequentially. In this context, we found the Laplace approximation to be reliable and numerically stable, because the lack of data resulted in a regularization effect.
We also implemented the modified double-loop EP algorithm from \cite{jylanki2011robust}, but preliminary results resulted in poor performance with many iterations converging to the wrong solution or not converging at all. We conjecture that this effect is because of the limited data available, and further research is required. For brevity, we present results in \secref{sec:experiments} using only the Laplace approximation.

We are interested in computing the predictive posterior from equation \eqref{eq:predgp} with the new likelihood function. The Laplace approximation for the conditional posterior  of the latent function, which we write as $p(f | \y, \X, \Lambda, \sigma_0^2, \nu)$, is constructed from the second order Taylor expansion of log posterior around the mode $\hat{f}$, which results in a Gaussian approximation:
\begin{equation}
  \label{eq:postst}
  p(f | \y, \X, \Lambda, \sigma_0^2, \nu) \approx \mathcal{N}(f | \hat{f}, \Sigma),
\end{equation}
where $\hat{f} = \arg\max p(f | \y, \X, \Lambda, \sigma_0^2, \nu)$ is the maximum \textit{a posteriori} and $\Sigma^{-1} = \K^{-1} + \mathbf{W}$ the Hessian of the negative log conditional posterior at the mode with, $\mathbf{W} = diag_i\left(\nabla_{f_i} \nabla_{f_i} \log p(y|f_i, \sigma, \nu)|_{f_i=\hat{f_i}}\right)$.

Finally, the new predictive distribution can be computed by marginalization of equation \eqref{eq:postst}. That is:
\begin{equation}
  \label{eq:predtlik}
  \begin{split}
    \mu(\x_q) &= \kv^T\K^{-1}\hat{f}, \\
    \sigma ^2 (\x_q) &= k - \kv^T \left(\K + \mathbf{W}^{-1}\right)^{-1} \kv,
  \end{split}
\end{equation}
where $k = k(\x_q, \x_q)$ and $\kv = \kv(\x_q,\X)$.
We refer to Vanhatalo et al. \cite{vanhatalo2009student} for implementation details.

\subsection{Student-$t$ process with additive kernel noise\label{sec:tprocess}}
For comparison, we also include the Student-$t$ process from Shah et al. \cite{AmarShah2014}, which we will the define in terms of the conditional posterior in the form of a multivariate Student-$t$ distribution. This process completely changes the model presented in \secref{sec:bayesianoptimization}. For brevity we do not include the equations for the predictive distribution, hyperparmenter optimization and expected improvement with the new model. These can be found in the literature \cite{AmarShah2014,martinez2014bayesopt,Santner03,Williams_Santner_Notz_2000}.

In this case, the Student-$t$ process is generated by placing an inverse gamma prior\footnote{Note that the inverse gamma is also equivalent to the scaled inverse $\chi^2$ distribution $\chi^{-2}_{\nu} (\sigma^2_0) = \IG\left(\frac{\nu}{2}, \frac{\nu\sigma^2_0}{2}\right)$, which will define the Student-$t$ in terms of the degrees of freedom \cite{O'Hagan1992} as mentioned in \secref{sec:robust}.} on the scale parameter of the kernel matrix \cite{martinez2014bayesopt}, that is, at the stage we replace the kernel matrix from equation \eqref{eq:covmatrix} to
\[
  \K = \sigma_s^2 \left[\begin{pmatrix}\kv(\x_1,\X) &\ldots& \kv(\x_t,\X)\end{pmatrix} + \mathbf{I}\sigma^2_n\right]
\]
 with $\sigma_s^2 \sim \IG(a,b)$. This is the multivariate generalization of equation \eqref{eq:likthierarchical}. Note also how the signal $\sigma_s^2$ and noise $\sigma_n^2$ variances become entangled. As reported by Shah et al. \cite{AmarShah2014}, this results are analogous to the method of using the inverse Wishart process as a prior on $\K$. The multivariate Student-$t$ distribution that generate the corresponding process is defined as:
\begin{equation}
  \label{eq:logliktrass}
  \begin{split}
  t(\y;m,\Sigmab,a,b) =& \frac{\Gamma\left(a + \frac{n}{2}\right)}{\Gamma(a)}\frac{1}{\sqrt{(2a \pi)^{n}\left|b^{-1}\K\right|}}\\&\left[1+\frac{b(\y-m)^T\K^{-1}(\y-m)}{2a}\right]^{-a - \frac{n}{2}}    
  \end{split}
  \end{equation}
where $m=m(\x)$ is the mean function, which is generally assumed to be $m(\x) = 0$ and $a$ and $b$ are the parameters of the inverse gamma. Again, we refer to the literature for implementation details about the posterior inference \cite{AmarShah2014,martinez2014bayesopt,Santner03,Williams_Santner_Notz_2000}.

\section{OUTLIER DIAGNOSTICS\label{sec:diagnostics}}
This part of our method is independent of the robust regression model selected before (see \secref{sec:laplace} and \secref{sec:tprocess}), although for clarity we will assume that we are using the GP with Student-$t$ likelihood from \secref{sec:laplace}. Once we have built the robust regression model, we are able to identify the outliers from the rest of the data. As can be seen in \figref{fig:example}, the mean function computed with the robust regression (center) is not biased like the nonrobust regression (left). Therefore, we can determine that the outliers are the points in the tail of the predictive distribution. For example, note how the point close to $(-2, 2)$ introduces a large bias in the nonrobust regression. As a result, in the nonrobust regression, the mean prediction is much closer to the point.

For that purpose, we compute the upper and lower $\alpha$-percentile of the predicted distribution as a classification threshold, where $\alpha$ is the assumed level of outliers. The selection of this parameter will determine the number of false positives and false negatives. High values of $\alpha$ will classify many points as outliers, reducing the effective sample size for Bayesian optimization. On the other hand, low values of $\alpha$ reduce the robustness of the method by misclassifying actual outliers.

\paragraph{No permanent rejection}
In theory, assuming that a single observation arrives per iteration, only that last observation should be questioned.
However, because new data helps improve the model, we found that reclassifying all the points worked better, as new information allows better classification over past observations. Sometimes, points that initially were considered outliers can be found part of the model while, more frequently, points that were initially misclassified as acceptable are properly detected with a better model. For Bayesian optimization, the general assumption is that data points are \emph{expensive} in some sense (such as time or energy), thus no point is permanently deleted or ignored.

\paragraph{Scheduling diagnostics}
Although the Student-$t$ likelihood is able to identify $m$ outliers out of $2m$ points, we have found that in practices it is reasonable to wait for a certain number of iterations before starting classifying data. The motivation is to have a proper regression model with a correct estimate of the hyperparameters. We also found that, because of the sequential nature of Bayesian optimization, if the last point is misclassified as an outlier and removed in the Bayesian optimization, there is a large probability that the will be selected again in the next iteration, which will again might result in a misclassification and so on, wasting valuable resources. Finally, the computational cost of the Student-$t$ likelihood is more expensive than the Gaussian likelihood. Therefore, we propose to use the Student-$t$ likelihood and posterior data filtering after $\ninit$ points and, then only once out of each $n_s$ subsequent iterations.

Finally, once the outliers are classified and removed, the optimization is performed with a standard GP computed only with the remaining points, because it produces more stable and fast solutions (see \figref{fig:example}). This proved especially true at early stages, when the regression model is noisy and inaccurate, and some large misclassifications might happen. Knowing that there is a limitation on the number of outliers that the Student-$t$ distribution is robust, we are able to detect if there has been a failure in the filtering process by checking if the number of outliers is larger than $m$ for a total of $2m$ points.

\begin{algorithm}
\renewcommand{\algorithmicrequire}{\textbf{Input:}}
\caption{BO with outliers\label{al:robustbo}}
\begin{algorithmic}[1]
  \Require Total budget $T$, rejection threshold $\alpha$ 
  \State Initial design of $p$ points (e.g.: LHS)
   \Statex \hspace{\algorithmicindent} $\X~\gets~\x_{1:p} \qquad  \y \gets y_{1:p}$
\Statex
\For{$t = p+1 \ldots T$}
 \If {schedule($t$)}
   \State $\mathbf{\Theta_t} \gets \text{fitGPwithTlik}(\X,\y)$
   \State $\X_{in}, \y_{in} \gets \text{filterOutliers}(\X,\y,\mathbf{\Theta_t}, \alpha)$
 \EndIf
   \Statex
   \If {$length(\y_{in}) < \lfloor length(\y) / 2 \rfloor$ \textbf{or} \\\hspace{\algorithmicindent} \textbf{not} schedule($t$)}
   \State $\X_{in} \gets \X \qquad  \y_{in} \gets \y$
   \EndIf
   \Statex
   \State $\mathbf{\Theta_g} \gets \text{fitGPwithGlik}(\X_{in}, \y_{in})$ 
   \State $\x_t = \arg\max_{\x} \; EI(\x | \X_{in}, \y_{in},\mathbf{\Theta_g})$
   \State $y_t \gets f(\x_t) \qquad \X \gets add(\x_{t}) \qquad  \y \gets add(y_{t})$
\EndFor
\end{algorithmic}
\end{algorithm}

\section{BAYESIAN OPTIMIZATION WITH OUTLIERS\label{sec:algorithmdefinition}}

Our method is summarized in \algoref{al:robustbo}. For those familiar with Bayesian optimization will recognize lines 1, 2 and 9-11 as the standard procedure: draw some initial points and select each new point based on the expected improvement computed using a fitted GP. As pointed out before, our contribution uses elements from the robust regression literature and the outlier diagnostics. Again, at this point, we assume that we are going to use the robust regression method from \secref{sec:laplace} which we have represented as the function \texttt{fitGPwithTlik}, although the algorithm would work with the Student-$t$ process from \secref{sec:tprocess} by replacing the function.

Next, we filter the outliers with the function \texttt{filterOutliers} introduced in \secref{sec:diagnostics}. For this purpose, we need a predefined $\alpha$ term which defines the part of the tail that belongs to outliers. For our results, we have used $\alpha = 0.05$ which correspond to the $5\%$ percentile. We also considered lower values, such as, $\alpha = 0.01$ which reduced the number of false positives. We found than an agressive threshold worked better in practice because we are not permanently rejecting any point. False positives are usually correctly classified in subsequent iterations, when more data is available. Furthermore, as commented in \secref{sec:diagnostics}, we know that for the Student-$t$ robust, we need at least half of the points to be inliers. Therefore, we can monitor at every iteration if the number of inliers \texttt{length}($\y_{in}$) is at least half of the total number of points \texttt{length}($\y$) .

Finally, for the \texttt{schedule}, in most experiments we have used an initial delay of $\ninit = 10$ iterations and the filtering was performed one every $n=2$ iterations. However, we found that the results were not fundamentally different using different schedules and it is more of a tradeoff of the computation cost, convergence speed and total budget. For example, in the experimental results we used a different schedule in the robot walker experiment from \secref{sec:walker}.

\section{RESULTS\label{sec:experiments}}

We evaluated our method on a set of benchmarks and realistic applications. For the benchmarks we have compared a set of different methods: \emph{our method} according to algorithm \algoref{al:robustbo}; BO with \emph{$t$-likelihood} and \emph{$t$-process} which corresponds to the methods presented in \secref{sec:laplace} and \secref{sec:tprocess} respectively where all the outliers are accommodated in the regression model and no rejection step is performed; \emph{baseline} which uses the standard Gaussian likelihood as presented in \secref{sec:gp}. We also included a \emph{no outliers} experiment which corresponds to the ideal scenario with no outliers present. For reproducibility purpose, the outliers were artificially generated in all cases so that the distribution is equivalent for all the methods. Common random numbers were used between different methods. All the experiments were repeated for 10\% and 20\% outlier proportion using exactly the same configuration parameters to illustrate the robustness of the application to different levels of outliers. All plots represent the average outcome and 95\% confidence bounds over 20 trials.

\begin{figure}
	\centering
	\includegraphics[width=0.9\linewidth]{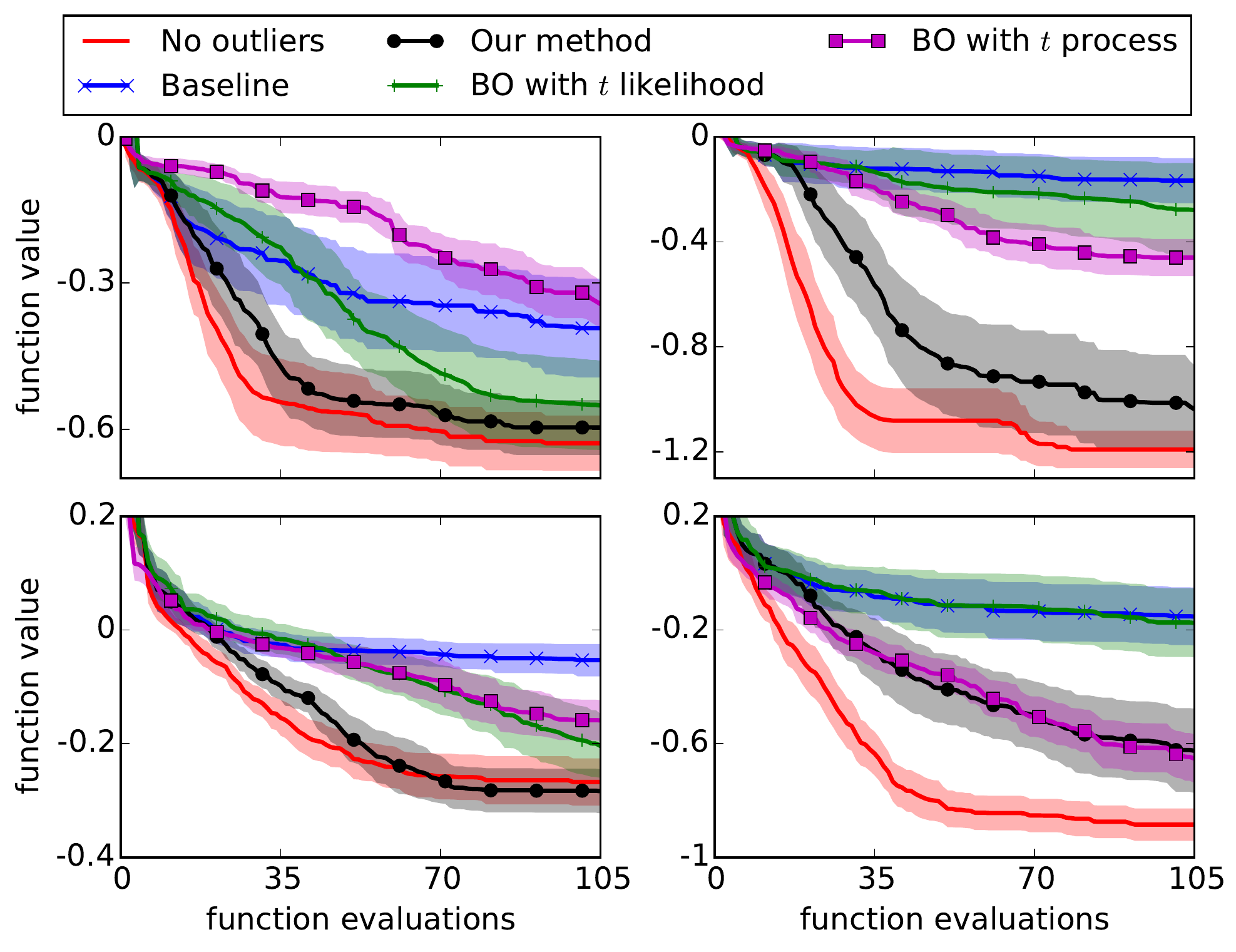}
	\caption{
		8D randomly generated functions with outliers with various Bayesian Optimization strategies.
		\textit{left}: 10\% outliers.
		\textit{right}: 20\% outliers.
		\textit{top}: Mat\'ern generated (within model comparison).
		\textit{bottom}: rational quadratic generated (out-of-model comparison).
		\label{fig:matplot}
	}
\end{figure}

\subsection{Numerical benchmarks}

For the numerical benchmarks we used the methodology from Henning and Schuler \cite{HennigSchuler2012}. We  generated a set of 8D random functions from two types of Gaussian processes. For the \emph{within model comparison}, we have generated the samples from a GP with the same $C^4$ Mat\'ern kernel used in optimization; while for the \emph{out-of-model} comparison, we generated the samples from a GP with a rational quadratic kernel with $\alpha=2$. Outliers were iid sampled from a uniform distribution $y_{outlier} \sim \mathcal{U}(1,2)$. Results can be seen in \figref{fig:matplot}. For each experiment configuration (kernel and ratio of outliers), we generate a different random function, resulting in different ranges for the vertical axis. Our method outperforms both robust regression methods and it is able to reach performance comparable to not having outliers at all. We can also see that if not considering the outliers, the effect is devastating, resulting in the optimization being stuck. Between the robust regression methods there is no clear winner. We found the Student-$t$ likelihood to be more reliable between experiments, thus we decided to use that approach within our own method as pointed out in \secref{sec:algorithmdefinition}.

\begin{figure}
\centering
    \includegraphics[width=0.9\linewidth]{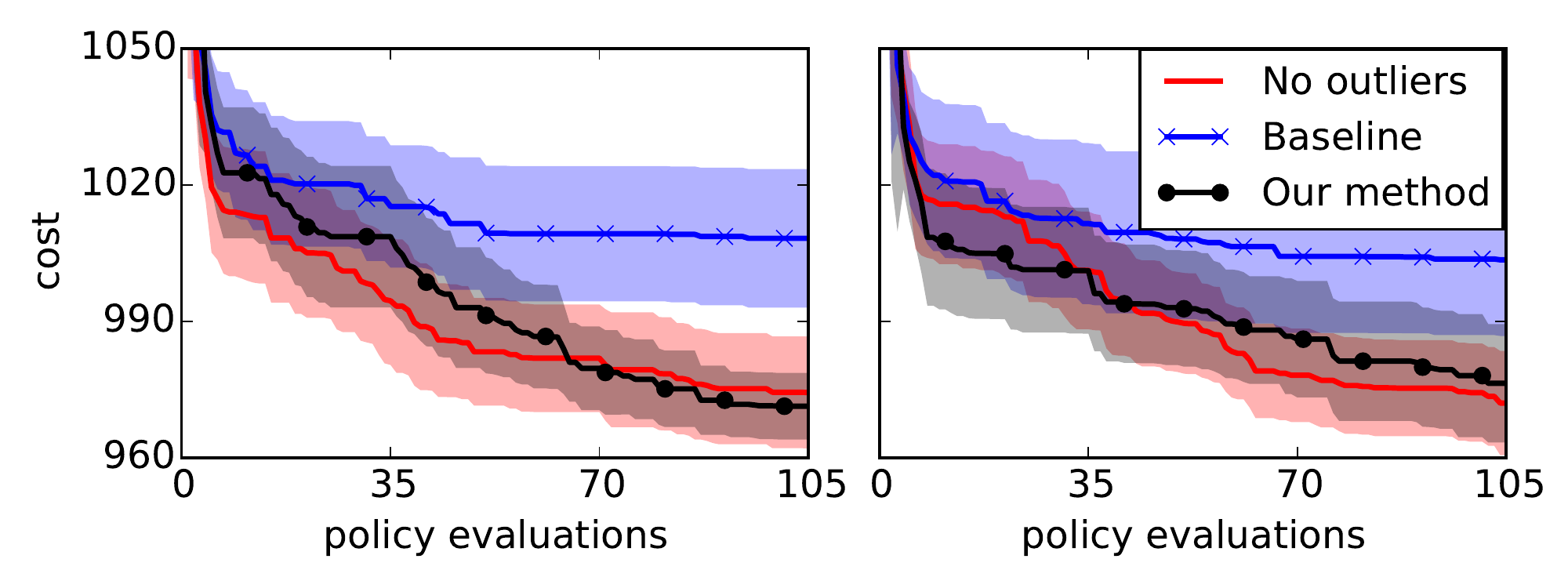}
	\caption{
		Optimization of the robot walking policy.
		\textit{left}: For a 10\% outlier rate, the Student-$t$ likelihood is able to prune some of the out-of-model points which allows better refinement than the standard GP baseline.
		\textit{right}: When the number of outliers is larger (20\%), the Student-$t$ likelihood allows us to roughly recover the performance in the absence of outliers.
		\label{fig:robotplot}
	}
\end{figure}

\begin{figure}
	\centering
    \includegraphics[width=0.9\linewidth]{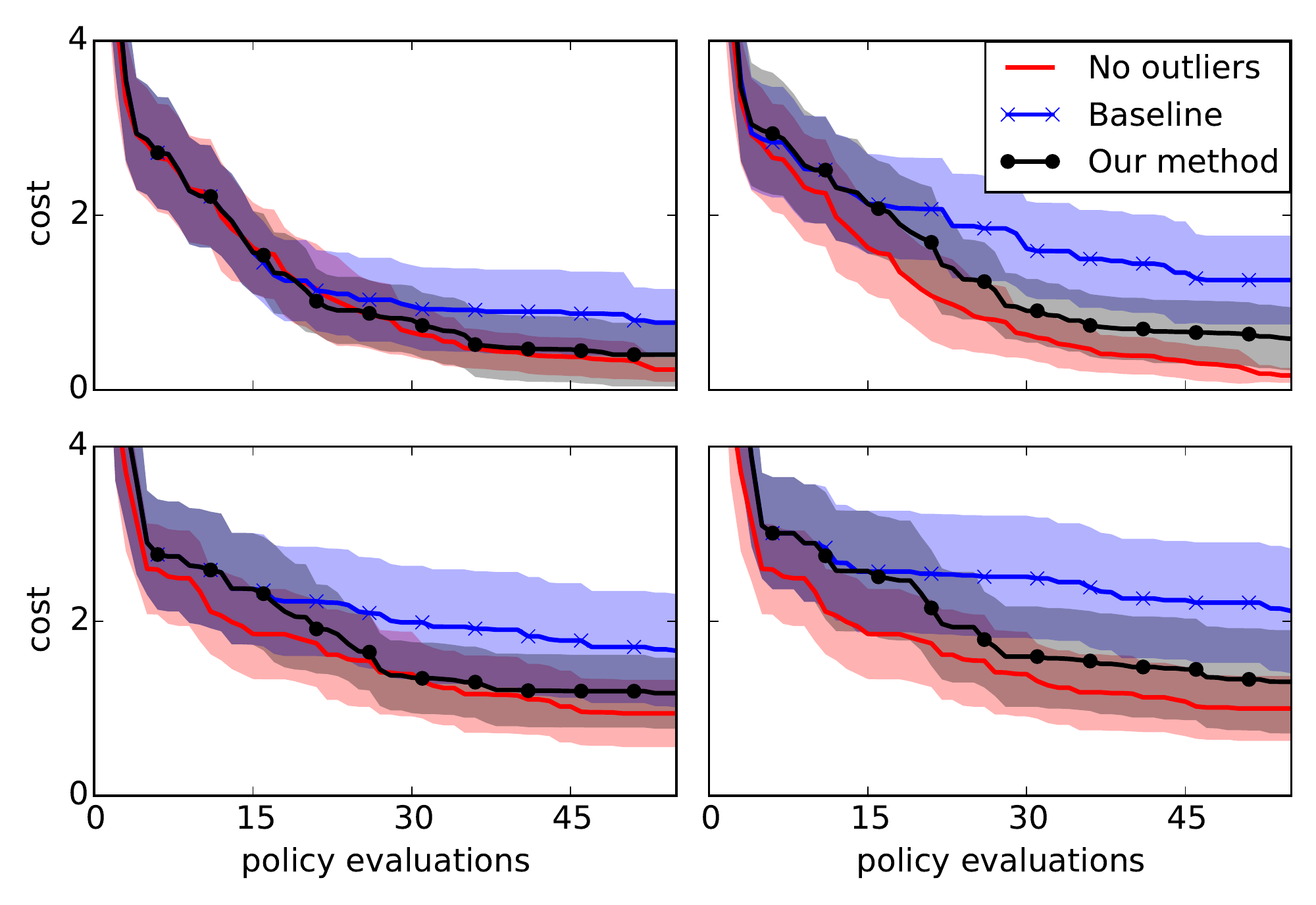}
	\caption{
		Optimization of the robot pushing policy.
		\textit{top left}: 3D input, 10\% outliers.
		\textit{top right}: 3D input, 20\% outliers.
		\textit{bottom left}: 4D input, 10\% outliers.
		\textit{bottom right}: 4D input
                , 20\% outliers.
		\label{fig:robotpush}
	}
\end{figure}

\subsection{Robot planning and control\label{sec:walker}}

Active policy search \cite{MartinezCantin07RSS} is a reinforcement learning method to control a robot or autonomous agent by refining its policy using Bayesian optimization on the reward function. It has been successfully applied for robot walking in controlled environments \cite{Calandra2015a}. In this case, the objective of the optimization is to find a stable policy, even in the presence of external perturbations. However, in some trials, the robot might find obstacles or perturbations that are physically impossible to overcome. Thus, the robot returns a poor reward even if the policy is good in other conditions.

For example, these days it is common to find experiments of robot learning walking patterns in laboratory conditions, where external interferences are reduced or controlled. In many cases, the objective of the learned controller is to be able to react to some of those external perturbations, like a light push or a terrain slope. However, in the near future, robots will have to learn and adapt in all kind of environments with uncontrolled conditions, some of which would be physically impossible to compensate. Thus, the learning process must be able to identify when the failure is due to a bad policy or a strong perturbation.

\paragraph{Robot walking} For this experiment, we have used a full body dynamic simulator \cite{robotsimulator} 
of a robot walking along with a predefined set of controllers from which we selected 6 parameters to tune, the stance and swing acceleration terms. In the scenarios with outliers, we have simulated the failures as the robot reaching a insurmountable obstacle at a random time during the trajectory, resulting in the robot tripping and falling. Therefore, the resulting reward is similar to the reward obtained with a bad policy, which also results in a \emph{crash} state at different times. \figref{fig:robotplot} shows the results across 30 trials.

It has been shown that robot policy search is a complex problem for Bayesian optimization due to the non-stationary behavior of many reward functions \cite{MartinezCantin17icra}. A large number of nearly flat \emph{crash} results yield combined with large variability near the optimum results in an underperforming GP model because all the results cannot agree to a single stationary function. In this case, the Student-$t$ likelihood is also classifying some of the actual \emph{crash} states as outliers because they do not agree with the regression model, resulting in a subtle improvement over no having outliers.

\begin{figure*}
	\centering
	\includegraphics[width=0.7\linewidth]{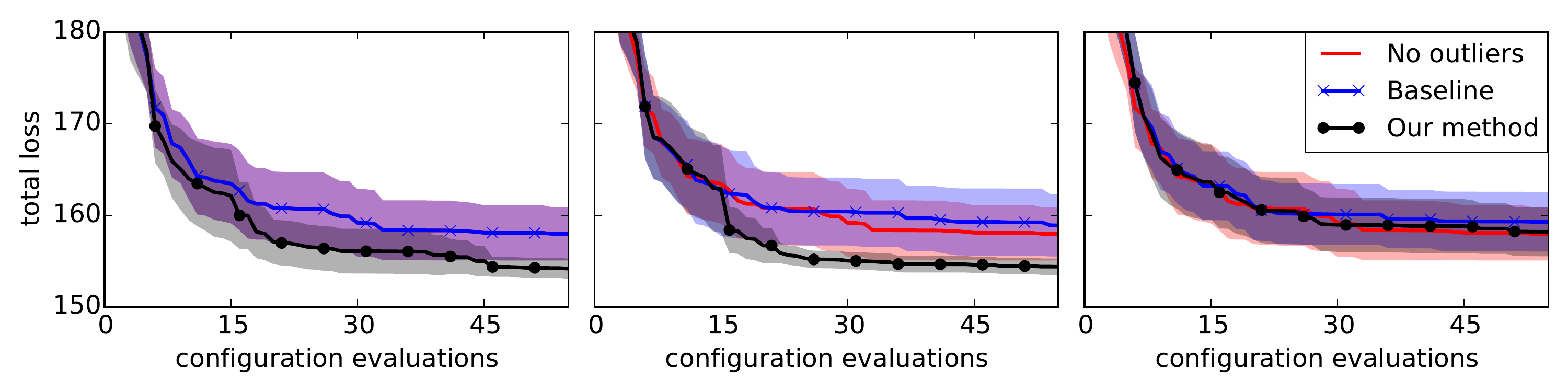}
	\caption{
		Optimization of the variational autoencoder on the MNIST dataset.
		\textit{left}: No outliers (thus the ``Baseline'' coincides with the ``No outliers case'').
		\textit{center}: 10\% outliers.
		\textit{right}: 20\% outliers.
		\label{fig:vae}
	}
\end{figure*}

\paragraph{Robot pushing} We have replicated the experiments from Wang and Jegelka \cite{pmlr-v70-wang17e} for robot pushing. The experiment is based on the \emph{pre-image} setup from Kaelbling and Lozano-Perez \cite{LPKTLPICRA16} and consist on performing active policy search to select the pushing action that minimize the distance of the pushed object to the goal location. The objective is to find a good pre-image for pushing the object to the designated goal location. The first function we tested has a 3-dimensional input: robot location $(r_x , r_y )$ and pushing duration $t_r$. The second function has a 4-dimensional input: robot location and angle $(r_x, r_y, r_{\theta})$, and pushing duration $t_r$. We select 20 random goal locations for each function to test if our method can learn where to push for these locations. In normal conditions, the goal was placed in a reachable position. Failures and outliers were modeled by placing the object just outside the reachable region to represent a configuration or sensor problem for which the distance to the goal is incorrectly measured.  \figref{fig:robotpush} shows the results of both the 3D and 4D problems.

\begin{figure}
	\centering
	\includegraphics[width=0.55\linewidth]{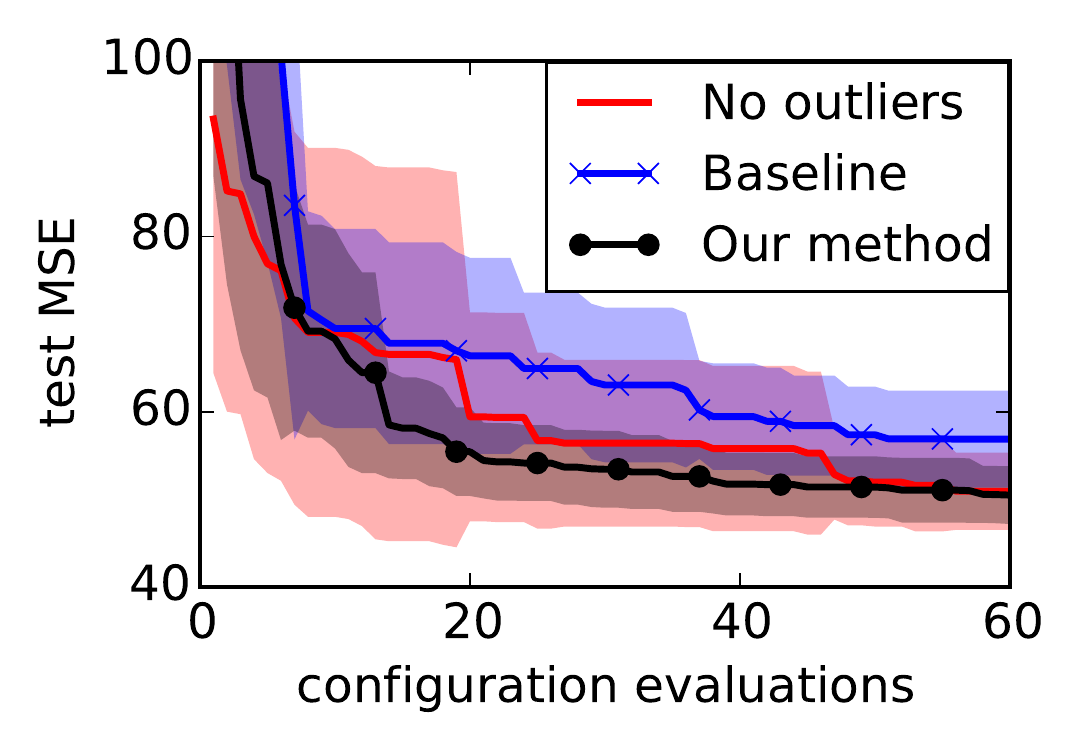}
	\caption{
		Optimization of the feed-forward neural network on the Boston housing dataset.
		\label{fig:boston}
	}
\end{figure}

\subsection{Hyperparameter tuning}

For the problem of hyperparameter tuning for which Bayesian optimization is growing in popularity, there are many possibilities for outliers: bugs, failures, etc. There are some sources of outliers that are intrinsic to the procedure. For example, during hyperparameter tuning, early stopping during training might be pushed to the limit to guarantee no overfitting and reduce the already expensive computational cost. There are methods which directly reallocate resources based on the performance at early stages \cite{klein-iclr17}. On the other hand, the initialization is nontrivial \cite{pmlr-v28-sutskever13}, with random initialization of variables resulting in very different behaviors at early stages. For instance, the performance of a set of hyperparameters may seem poor after few epochs because the initialization occurred in a complex region (flat, saddle points, etc.) and not because the set of hyperparameters was worse than others. Therefore, early stopping results in some points being actual outliers. This effect is present in our experiments, where our method is able to achieve better performance than standard BO with no induced outliers.

\paragraph{Variational autoencoder} Variational autoenconders (VAE) are a powerful generative method for deep learning. In this experiment we train a VAE for the MNIST dataset \cite{kerasVAE}.
The hyperparameters we tune are the number of nodes in the hidden layer and learning rate, learning rate decay, and $\epsilon$ constant for the Adam optimizer. In this case, an outlier simulates an IO failure where the VAE is trained only on a subset of the data (randomly generated between 100 and 1000 images). The results are shown in \figref{fig:vae}. We can see how even in the case were no outliers were induced externally, our method outperforms standard BO, which reinforces the theory that there are already outliers present. Besides, for a high level of outliers (20\%) the performance drop, suggesting that the number of simulated outliers in addition with the existing outliers reaches a point near the limit of robustness.

\paragraph{Feedforward network} Inspired by Wang and Jegelka~\cite{pmlr-v70-wang17e}, we use a single layer feedforward neural network on the Boston housing dataset. The hyperparameters we tuned were the number of nodes in the hidden layer, learning rate, learning rate decay, and $\rho$, the parameter that controls the exponential decay rate from RMSprop. \figref{fig:boston} shows the results of the optimization for 10\% outliers. An outlier in this optimization consisted of running the neural network $5$ epochs, rather than the standard $20$ epochs. We can see how the behaviour is similar to the VAE, improving the convergence over the case with no outliers.

\section{CONCLUSIONS}

We have presented a method to extend Bayesian optimization in the presence of outliers. The method combines robust regression with a Student-$t$ likelihood on a GP, and outlier analysis to classify inlier and outlier data points. We have extensively evaluated the proposed method in many benchmarks and realistic applications showing that our method is suitable for practical Bayesian optimization in the presence of outliers. Furthermore, we have seen how our method has been able to outperform standard Bayesian optimization in a controlled environment without induced outliers. This highlights the importance of this approach and the possible presence of outlier data even in supposedly controlled environments and lab conditions. Finally, we have experimentally proven that Bayesian optimization with a robust surrogate model designed to accomodate outliers produces suboptimal results.

\bibliographystyle{plain}

\end{document}